\documentclass[lettersize,journal]{IEEEtran}
\usepackage{amsmath,amsfonts}
\usepackage{algorithmicx}
\usepackage{algpseudocode}
\usepackage{algorithm}
\usepackage{array}
\usepackage[caption=false,font=normalsize,labelfont=sf,textfont=sf]{subfig}
\usepackage{textcomp}
\usepackage{stfloats}
\usepackage{url}
\usepackage{verbatim}
\usepackage{multirow}
\usepackage{graphicx}
\usepackage{booktabs} 
\usepackage{cite}
\usepackage[table]{xcolor}
\usepackage{soul} % 导入 soul 包
\usepackage{color, xcolor}
\usepackage{algorithm}
\usepackage{algorithmicx}
\usepackage{algpseudocode}
\usepackage{amsmath}
\usepackage{hyperref}

\begin{document}

\title{Accurate and Complete Surface Reconstruction \\ from 3D Gaussians via  Direct SDF Learning}

\author{Wenzhi Guo, Guangchi Fang, Mingqing Wei, Jie Qin, Bing Wang$^{\ast}$
\thanks{W. Guo is with the Department of Aeronautical and Aviation Engineering, The Hong Kong Polytechnic University, Hong Kong SAR, China and the Department of Computer Science and Technology, Nanjing University, Nanjing, China (e-mail: wenzhi.guo@connect.polyu.hk).}
\thanks{Mingqing Wei and Jie Qin are with School of Artificial Intelligence, Nanjing University of Aeronautics and Astronautics, Nanjing, China, and the  (e-mail: mqwei@nuaa.edu.cn, jie.qin@nuaa.edu.cn).}% <-this % stops a space
\thanks{Guangchi Fang and Bing Wang are with the Department of Aeronautical and Aviation Engineering, The Hong Kong Polytechnic University, Hong Kong SAR, China (e-mail: guangchi.fang@gmail.com, bingwang@polyu.edu.hk)}
}

\markboth{Journal of \LaTeX\ Class Files,~Vol.~14, No.~8, August~2021}%
{Shell \MakeLowercase{\textit{et al.}}: A Sample Article Using IEEEtran.cls for IEEE Journals}

% \IEEEpubid{0000--0000/00$00.00~\copyright~2021 IEEE}

\maketitle

\begin{abstract}
% 3D Gaussian Splatting (3DGS) has recently emerged as a powerful paradigm for photorealistic view synthesis through the use of spatially distributed Gaussian primitives. Despite its impressive rendering capabilities, accurate and complete surface reconstruction remains an open challenge due to the representation's inherently unstructured nature and lack of explicit geometric supervision. In this paper, we present \textbf{DiGS}, a novel framework that tightly integrates Signed Distance Field (SDF) learning into the 3DGS pipeline to impose strong and interpretable surface priors. By associating each Gaussian with a learnable SDF value, DiGS facilitates explicit surface alignment and enhances geometric consistency across views. To overcome the limitations of incomplete and non-uniform scene coverage, we introduce a depth-aware and normal-aligned grid growth strategy that adaptively guides the spatial distribution of Gaussians based on geometric cues. Extensive experiments across diverse benchmarks demonstrate that DiGS achieves superior surface reconstruction fidelity and completeness, while maintaining real-time rendering quality. Our results establish a new state-of-the-art in 3DGS-based surface reconstruction.
3D Gaussian Splatting (3DGS) has recently emerged as a powerful paradigm for photorealistic view synthesis, representing scenes with spatially distributed Gaussian primitives. While highly effective for rendering, achieving accurate and complete surface reconstruction remains challenging due to the unstructured nature of the representation and the absence of explicit geometric supervision. In this work, we propose \textbf{DiGS}, a unified framework that embeds Signed Distance Field (SDF) learning directly into the 3DGS pipeline, thereby enforcing strong and interpretable surface priors. By associating each Gaussian with a learnable SDF value, DiGS explicitly aligns primitives with underlying geometry and improves cross-view consistency. To further ensure dense and coherent coverage, we design a geometry-guided grid growth strategy that adaptively distributes Gaussians along geometry-consistent regions under a multi-scale hierarchy. Extensive experiments on standard benchmarks, including DTU, Mip-NeRF 360, and Tanks \& Temples, demonstrate that DiGS consistently improves reconstruction accuracy and completeness while retaining high rendering fidelity. Our code is available at \href{https://github.com/DARYL-GWZ/DIGS}{https://github.com/DARYL-GWZ/DIGS.}

\end{abstract}


% % 
\section{Introduction}

Recent advances in \textit{3D Gaussian Splatting} (3DGS) \cite{kerbl20233d} have transformed neural scene representation and rendering. By modeling a scene as a collection of anisotropic Gaussian primitives and rendering them through a carefully engineered rasterization pipeline, 3DGS achieves real-time, photo-realistic view synthesis. Its efficiency and fidelity have quickly spurred adoption in diverse applications, ranging from augmented and virtual reality \cite{zhai2025splatloc,song2024toward,jiang2024vr,lian2024integration} to autonomous driving \cite{zhou2024drivinggaussian,khan2024autosplat,hess2024splatad}.  

Despite these advances, accurate geometry reconstruction remains a fundamental challenge for Gaussian-based representations. Unlike volumetric fields or mesh-based methods, Gaussian primitives are inherently unstructured and provide only weak geometric cues under multi-view supervision. As a result, they often exhibit drift and misalignment with underlying surfaces, leading to limited reconstruction fidelity \cite{fei20243d,cheng2024gaussianpro,wu2024recent}. Moreover, existing growth strategies for Gaussian models \cite{kerbl20233d, yu2024gaussian} typically produce sparse and uneven distributions, failing to guarantee comprehensive scene coverage in complex or low-texture environments. These shortcomings suggest that current pipelines are largely appearance-driven, with geometry treated as an auxiliary objective rather than a core principle.

To address this, recent advances have sought to augment Gaussian splatting with stronger geometric priors. Geometric regularization methods \cite{huang20242d, chen2024pgsr} introduce constraints such as normals or silhouettes, while surface extraction pipelines \cite{guedon2024sugar,yu2024gaussian} recover meshes only in a post-processing stage. SDF-based formulations provide a more direct coupling to geometry: 3DGSR \cite{lyu20243dgsr} supervises Gaussians with signed distance values but suffers from artifacts under sparse guidance, and GSDF \cite{yu2024gsdf} adopts a dual-branch optimization that separates rendering and reconstruction, thereby increasing model complexity and weakening consistency between appearance and geometry. Similarly, PGSR \cite{chen2024pgsr} leverages composite geometric losses but requires post-training refinement for completeness, while Octree-GS \cite{yu2024gaussian} employs hierarchical multi-scale expansion guided primarily by appearance, often resulting in uneven coverage. Collectively, these approaches demonstrate the potential of SDF-augmented Gaussians, yet they leave unresolved the fundamental question of how geometry and appearance can be unified within a single, efficient framework. Collectively, these approaches demonstrate the potential of SDF-augmented Gaussians, yet they leave unresolved the fundamental question: how can geometry and appearance be unified within a single Gaussian framework, such that signed distance priors and growth strategies are jointly leveraged to achieve both accurate and complete surface reconstruction?

\begin{figure}[!t]
\centering
\includegraphics[width=3.5in]{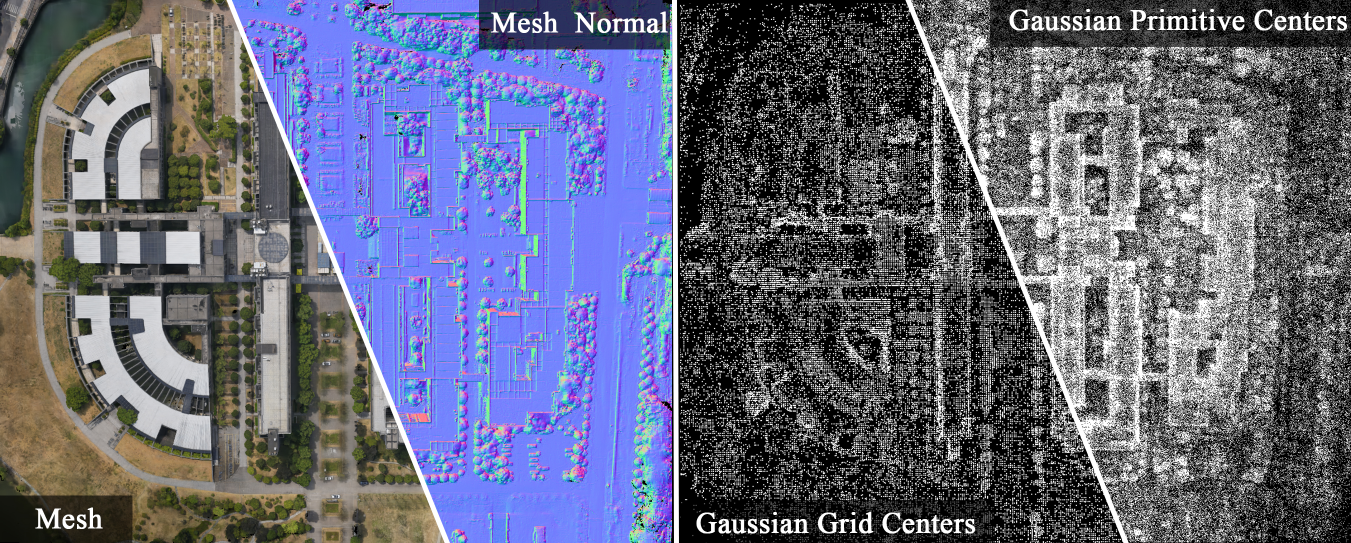}
\caption{\textbf{Conceptual Overview of DiGS.} DiGS advances Gaussian splatting from an appearance-driven representation toward a geometry-preserving paradigm, by embedding signed distance supervision into Gaussian primitives under a geometry-guided growth strategy.}
\label{top}
\end{figure}

In this paper, we present \textbf{DiGS}, a unified framework that advances Gaussian splatting from an appearance-driven representation to a geometry-preserving paradigm. Unlike prior approaches that either impose weak geometric regularization \cite{chen2024pgsr,guedon2024sugar} or decouple geometry and appearance into separate optimization branches \cite{lyu20243dgsr,yu2024gsdf}, DiGS integrates signed distance supervision directly into the Gaussian primitives, treating geometry as a first-class objective within the rendering pipeline. Each Gaussian is constrained by a learned signed distance value at its center, while a geometry-guided grid growth strategy systematically expands Gaussians along surface-consistent regions under a multi-scale level-of-detail (LoD) hierarchy. Crucially, SDF learning and geometry-guided grids growth are mutually dependent and jointly optimized: SDF supervision prevents Gaussian drift and grids growth guarantees surface completeness. Their tight coupling forms a necessary and self-reinforcing system, establishing DiGS as a new unified paradigm for geometry-aware Gaussian splatting that simultaneously achieves high-fidelity rendering and accurate surface reconstruction.

\begin{figure*}[!t]
\centering
\includegraphics[width=6.8in]{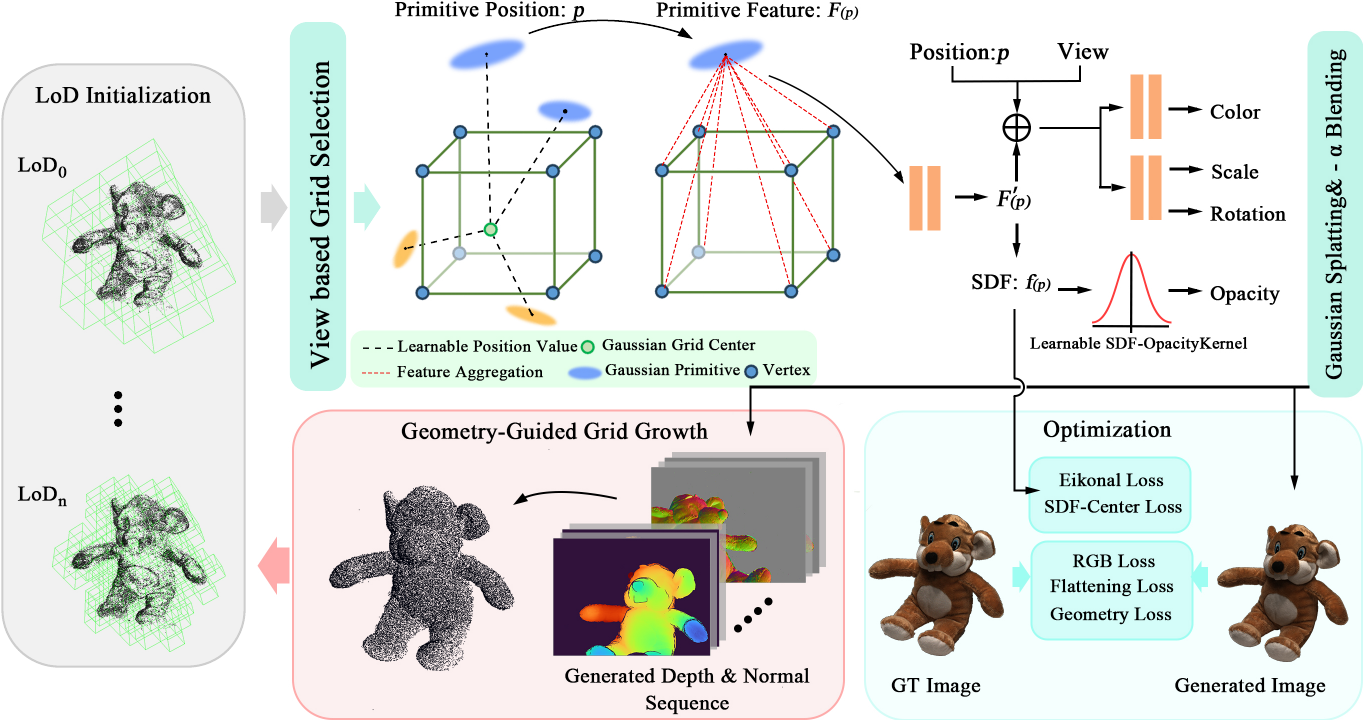}
\caption{\textbf{Detailed Framework of DiGS.} Starting from an SfM-initialized LoD grid, DiGS associates each cell with SDF-augmented Gaussians. Gaussian splatting and signed distance values are jointly predicted, with SDFs converted to opacity via a differentiable kernel. During training, a geometry-guided grid growth strategy expands Gaussians along geometry-consistent regions.}
\label{main}
\end{figure*}

The main contributions are summarized as follows:
\begin{itemize}
    \item We propose DiGS, a unified SDF-Gaussian framework, which directly embeds signed distance supervision into Gaussian primitives, achieving precise and consistent surface reconstruction.  
    \item We develop a geometry-guided Gaussian grid growth mechanism that facilitates dense and spatially coherent coverage, particularly effective for complex or low-texture geometries.   
    \item Extensive experiments on DTU, Mip-NeRF 360, and Tanks \& Temples demonstrate consistent improvements in reconstruction accuracy and completeness compared to state-of-the-art Gaussian-based and hybrid methods.  
\end{itemize}

\section{Related Work}

\subsection{Neural Surface Reconstruction}

Neural surface reconstruction has evolved as a result of the interplay between implicit representations and geometric constraints. The pioneering work of NeRF \cite{mildenhall2021nerf} introduced volumetric rendering for view synthesis, but it faced significant limitations in surface modeling due to its density-based formulation, which lacked explicit geometric constraints. To address this, NeuS \cite{wang2021neus} and VolSDF \cite{yariv2021volume} tightly coupled Signed Distance Fields (SDFs) with probabilistic volume rendering, enabling surface extraction through the zero-level set of the SDF, offering a more precise model for surface representation. These methods laid the foundation for surface-aware neural rendering but still struggled with handling complex geometries or large-scale scenes. A comprehensive survey of such neural implicit surface reconstruction paradigms is provided in \cite{zhang2025neural}, which systematizes recent advances and open challenges.

Subsequent works advanced the state of geometric fidelity through hybrid implicit-explicit representations. For example, UNISURF \cite{oechsle2021unisurf} unified implicit surfaces with volume rendering, offering improved detail preservation. Similarly, MonoSDF \cite{yu2022monosdf} and Geo-Neus \cite{fu2022geo} introduced monocular depth and normal priors for regularization, helping stabilize the optimization process by providing additional geometric priors. Extensions such as Vox-Surf \cite{li2022vox} explored voxel-based implicit surface representation, while H-SDF \cite{wang2024hsdf} proposed a hybrid sign-distance formulation to support arbitrary topologies. Indoor scene reconstruction has also benefited from hybrid priors, as shown in \cite{ye2024indoor}, where fine-grained geometry is recovered by leveraging normal enhancements. In parallel, semantic-driven surface modeling from point clouds, such as RangeUDF \cite{wang2022rangeudf}, demonstrated that semantic cues can further improve surface reconstruction robustness in challenging real-world environments.

Beyond hybrid priors, point-guided approaches such as PG-NeuS \cite{zhang2024pg} and its progressive-growing tri-plane extension PGT-NeuS \cite{xiang2025pgt} improved robustness and efficiency by leveraging sparse point correspondences and hierarchical representations. Prior-driven surface learning was further explored in PSDF \cite{su2024psdf}, which injects geometric priors into implicit learning for better multi-view reconstruction. In parallel, scalable formulations like \cite{yang2025scalable} addressed the challenge of high-quality reconstruction in large-scale settings, improving both accuracy and training efficiency. Complementary to these, decomposition-based formulations such as DM-NeRF \cite{wang2022dm} enabled geometry decomposition and manipulation directly from 2D images, highlighting the potential of scene factorization for enhancing surface-aware neural rendering.

Accelerated frameworks such as Instant-NGP \cite{muller2022instant} and Neuralangelo \cite{li2023neuralangelo} utilized multi-resolution hash grids and coarse-to-fine optimization strategies to enhance detail recovery and computational efficiency. Meanwhile, task-specific extensions emerged, for example H$_2$O-NeRF \cite{liu2025h} tailored radiance field reconstruction to the special case of two-hand-held objects, demonstrating the adaptability of neural implicit surfaces to constrained domains.

3DGSR \cite{lyu20243dgsr} was one of the first methods to integrate SDFs directly with 3D Gaussian Splatting (3DGS) to enforce surface constraints while maintaining rendering speed. This approach introduced loose coupling strategies to align Gaussian primitives with SDF-derived surfaces, achieving a more detailed reconstruction than previous density-based methods. Despite the notable progress, the scalability and computational efficiency of these methods remained a challenge, particularly for large-scale and high-resolution scenes. For instance, BakedSDF \cite{yariv2023bakedsdf} and NeRFMeshing \cite{rakotosaona2024nerfmeshing} require hours of training and struggle with efficiently handling large-scale scene reconstruction.

Recent advancements, such as H3DNet \cite{ramon2021h3d}, further emphasize the importance of geometric priors to stabilize the optimization of neural implicit surfaces, ensuring both geometric accuracy and robustness. Scaffold-GS \cite{lu2024scaffold} introduced hierarchical Gaussian structures for better alignment with scene geometry, but these approaches still face significant trade-offs between surface accuracy and computational efficiency.

\subsection{Gaussian Splatting Based Surface Reconstruction}

The introduction of 3D Gaussian Splatting (3DGS) \cite{kerbl20233d, fang2024mini,fang2024mini2}  marked a breakthrough in real-time rendering for neural surface reconstruction, by representing scenes as sets of anisotropic Gaussians that can be rasterized efficiently. While 3DGS excels in rendering photorealistic images, it inherently struggles with geometric accuracy due to the unstructured nature of Gaussian primitives. Early attempts to reconcile rendering efficiency with surface quality include methods like SuGaR \cite{guedon2024sugar}, which employed Poisson reconstruction, and 2DGS \cite{huang20242d}, which collapsed 3D Gaussians into planar surfels. These methods improved the overall quality but failed to preserve fine details or handle complex geometries.

NeuSG \cite{chen2023neusg} and GOF \cite{yu2024gaussian} explored hybrid implicit-explicit paradigms by using SDFs or opacity fields to guide the placement of Gaussians. However, they face critical limitations, particularly in the alignment between the Gaussian primitives and the underlying surface. For example, Scaffold-GS \cite{lu2024scaffold} improved the structural alignment of the Gaussians but sacrificed high-frequency details, while PGSR \cite{lu2024scaffold} used planar decomposition techniques that limited the adaptability of the approach to more complex topologies.

3DGSR \cite{lyu20243dgsr} proposed loose coupling strategies to align Gaussians with SDF-derived surfaces, enhancing the detail of the reconstruction while maintaining rendering efficiency. However, 3DGSR still faced challenges in achieving full geometric coverage and handling sparse data effectively.

Beyond these hybrid paradigms, several recent works expanded the scope and adaptability of 3DGS to various scenarios. MPGS \cite{li2025mpgs} introduced a multi-plane Gaussian representation to improve compact scene rendering, while LoopSparseGS \cite{bao2025loopsparsegs} proposed loop-based strategies for sparse-view friendly reconstruction. Fov-GS \cite{fan2025fov} leveraged foveated rendering to enable efficient handling of dynamic scenes, and RGAvatar \cite{fan2025rgavatar} extended Gaussian splatting to relightable 4D human avatar modeling. Similarly, PlGS \cite{wang2025plgs} addressed robust panoptic lifting, and Look-at-the-Sky \cite{wang2025look} specialized Gaussian splatting for outdoor environments. iVR-GS \cite{tang2025ivr} explored editable and explorable visualization via inverse volume rendering, and ARAP-GRF \cite{tong2025rigid} integrated deformation models for as-rigid-as-possible Gaussian field editing.

GSDF \cite{yu2024gsdf} later unified 3D Gaussian Splatting with SDFs through mutual geometry supervision, offering an improved surface representation. However, GSDF still struggled with computational efficiency and could not completely resolve issues related to large-scale scene reconstruction. GARF \cite{shi2022garf} introduced geometry-aware Gaussian refinement, but this method retained computational bottlenecks due to iterative density refinement processes.

These existing methods highlight the need for a framework that can tightly couple geometry and appearance while preserving the computational efficiency of 3D Gaussian Splatting. Our work in DiGS directly addresses these challenges by integrating SDF learning with 3D Gaussian Splatting in a unified manner. By combining normal-aligned growth and depth-guided grid expansion, DiGS ensures accurate surface reconstruction, robust spatial coverage, and high rendering efficiency. This innovative approach not only enhances geometric fidelity but also provides a scalable solution for  textureless scenes, addressing the limitations of previous methods.

\section{Preliminary}

\paragraph{\textbf{3D Gaussian Splatting}} 
3D Gaussian Splatting (3DGS) \cite{kerbl20233d} represents a scene as a collection of spatially distributed anisotropic 3D Gaussians \( \{ \mathcal{G}_i \}_{i=1}^N \), where each primitive is parameterized by its center \( \mathbf{p}_i \in \mathbb{R}^3 \), a covariance matrix \( \boldsymbol{\Sigma}_i \in \mathbb{R}^{3 \times 3} \), color \( \mathbf{c}_i \in \mathbb{R}^3 \), and opacity \( \alpha_i \in [0,1] \). The covariance matrix is typically factorized into a scaling and rotation component to define elliptical spatial support.

To render an image, each Gaussian is projected into screen space, and its contribution is accumulated via front-to-back \(\alpha\)-compositing:

\begin{equation}
\mathbf{C}(\mathbf{u}) = \sum_{i \in \mathcal{N}} T_i \alpha_i \mathbf{c}_i, \quad T_i = \prod_{j=1}^{i-1} (1 - \alpha_j),
\end{equation}
where \( \mathcal{N} \) denotes the sorted set of Gaussians affecting pixel \( \mathbf{u} \), and \( T_i \) represents the transmittance.

While 3DGS delivers photorealistic rendering with real-time performance, it lacks explicit geometric structure, and its opacity-driven formulation provides limited surface-awareness. Our work builds on this foundation by embedding signed distance priors directly into the Gaussian representation, enabling explicit surface alignment while preserving rendering efficiency.

\paragraph{\textbf{Surface Regularization}} 
To enhance geometric fidelity in multi-view reconstruction, recent methods introduce regularization objectives that enforce consistency across spatial and photometric domains. For instance, PGSR \cite{chen2024pgsr} proposes a composite geometric loss \( L_{\text{geo}} \) consisting of: (i) single-view geometric priors \( L_{\text{svgeo}} \), (ii) multi-view geometric consistency \( L_{\text{mvgeo}} \), and (iii) photometric alignment loss \( L_{\text{mvrgb}} \). These constraints improve robustness in untextured regions and guide reconstructions toward structurally plausible surfaces.

Instead of optimizing loss terms post hoc, our framework integrates geometric priors directly into the representation by learning signed distance values at each Gaussian center, inherently encoding surface proximity and enabling structure-aware supervision during rendering and growth.

\section{Method} 
We propose DiGS, a unified framework that tightly couples implicit surface modeling with explicit radiance representation under the Gaussian splatting paradigm. Rather than treating geometry and appearance as disjoint components, DiGS embeds a Signed Distance Field (SDF) directly into each Gaussian primitive, and integrates a multi-scale Level-of-Detail (LoD) structure with a geometry-guided growth strategy. In this way, surface priors are inherently encoded within the rendering process, leading to precise surface alignment and efficient, adaptive scene representation. Importantly, all components of our approach are co-designed and jointly optimized end-to-end. This is in stark contrast to prior methods like GSDF \cite{yu2024gsdf}, which separate geometry and radiance into dual branches, or Octree-GS \cite{ren2024octree} and PGSR \cite{chen2024pgsr}, which perform multi-resolution refinement as a post-process. By unifying these aspects in a single pipeline, our method ensures that geometry and appearance remain consistent and mutually reinforcing throughout optimization. We next detail each part of the method, organized as follows: an SDF-Gaussian representation (IV-A) defines our base scene parameterization, a geometry-guided grid growth mechanism (IV-B) progressively refines this representation, and a set of optimization objectives (IV-C) enforces geometric fidelity and photometric accuracy. 
\subsection{Unified SDF-Gaussian Representation} 
Our scene representation augments each Gaussian splat with an explicit SDF value and is structured in a multi-scale fashion. This SDF-Gaussian representation provides a common foundation for modeling both surface geometry and appearance. Unlike approaches that simply combine separate modules, we design this representation as a unified structure. In particular, the SDF at each Gaussian influences its rendering opacity, and the LoD hierarchy is built to accommodate the SDF-guided growth of Gaussians. 
\subsubsection{Octree-Based LoD Initialization} 
To efficiently cover scenes at multiple scales, we begin by partitioning space with a hierarchical octree grid. Each octree level $\ell=0,1,\dots,L_{\max}$ corresponds to a voxel size:
\begin{equation}
s_\ell = s_0 \cdot 2^\ell.
\end{equation}
\\
where $s_0$ is the base voxel size. Coarser levels (small $\ell$) thus span larger regions with fewer primitives, while finer levels (large $\ell$) focus on local details. We choose the number of levels $L_{\max}$ based on scene complexity and desired detail.\par
This ensures we capture fine details without an excessive number of primitives. The LoD structure enables adaptive refinement: broad areas are first represented coarsely, and only regions with complex geometry receive finer subdivisions. We initialize the octree using a sparse Structure-from-Motion (SfM) point cloud, which provides an initial approximation of scene geometry. Starting from the root (level 0) covering the entire scene, we subdivide octree cells down to a certain depth where sufficient SfM points fall inside. This yields an initial set of occupied voxels across levels, giving low-resolution coverage of the scene. Notably, the LoD grid in DiGS is not a fixed, static structure: it will be dynamically updated during training by our growth procedure . 
\begin{figure}[!t]
\centering
\includegraphics[width=3.5in]{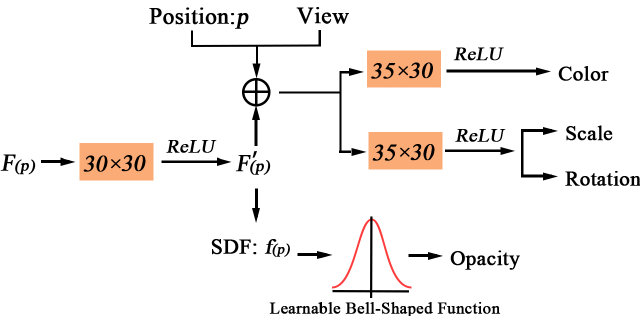}
\caption{Network Architecture. Given Gaussian position and view direction as inputs, an MLP predicts radiance features and signed distance values. The SDF is mapped to opacity through a learnable transfer function, ensuring surface awareness, while the radiance branch produces color for rendering.}
\label{net}
\vspace{-0.3cm}
\end{figure}
\subsubsection{Gaussian Position Encoding} Within each occupied octree cell, we allocate a small set of Gaussian primitives to represent local geometry and appearance. Specifically, each voxel at level~$\ell$ is associated with $k$ Gaussians (with $k$ is 10 in our implementation). We parameterize the positions of these Gaussians relative to the voxel center $\mathbf{x}_v$ as:
\begin{equation}
\left\{\boldsymbol{p}_0, \ldots, \boldsymbol{p}_{k-1}\right\} = \mathbf{x}_v +  \left\{\mathcal{C}_0\cdot\mathcal{L}_0, \ldots, \mathcal{C}_{k-1}\cdot\mathcal{L}_{k-1}\right\}.
\end{equation}

where each $C_i\in\mathbb{R}^3$ is a learnable offset and $L_i\in\mathbb{R}^3$ is a learnable scale factor. Intuitively, $C_i$ and $L_i$ allow each Gaussian to shift from the cell center and spread around the voxel. This neural position encoding grants the model flexibility to fit complex local surface geometry: by adjusting the offsets, multiple Gaussians in a cell can align along different surface structures. \par
Each Gaussian is also associated with additional properties, such as a color and an anisotropic covariance matrix that defines its spatial extent. This design is crucial for capturing high-frequency geometry in regions of interest while keeping less interesting regions coarse. It also naturally complements the LoD structure by allowing newly subdivided cells to inherit and refine the features from their parent cell. \par
At any 3D Gaussian primitive center $\mathbf{p}$, we can query the neural grid features via interpolation. We employ an inverse distance weighting (IDW) scheme to aggregate features from nearby voxel centers. Specifically, we find the eight surrounding grid vertices that form the corners of the cell containing $\mathbf{p}$ at the appropriate LoD level, and compute an interpolated feature $F(\mathbf{p})$ as:
\begin{equation}
F(p) = \sum_{i=1}^8 \frac{w_i}{\sum_j w_j} {f^{\prime}}, \quad \text{where } w_i = \frac{1}{\|\mathbf{x}_i - p\| + \epsilon}.
\end{equation}

where $f_i$ is the feature stored at vertex $\mathbf{x}_i$, and $\epsilon$ is a small constant for numerical stability. This weighted sum smoothly blends the features of the nearby Gaussians or grid nodes, ensuring that $F(\mathbf{p})$ changes continuously as $\mathbf{p}$ moves through space. We then decode $F^\prime(p)$ from $F(p)$ via MLP and directly split the one-dimensional value of the decoded neural Gaussian Primitive feature $F^\prime(p)$ as the SDF value $f(p)$, the whole process is shown on Figure \ref{net}. \par
\subsubsection{SDF Supervision and Opacity Modulation} 
To infuse geometric awareness into each Gaussian, we assign a Signed Distance value to its center. Let $f(\mathbf{p})$ denote the signed distance from point $\mathbf{p}$ to the nearest surface in the scene, with the convention that $f(\mathbf{p})=0$ on the surface, $f(\mathbf{p})>0$ outside (in free space), and $f(\mathbf{p})<0$ inside solid regions. In our representation, each Gaussian center $\mathbf{p}_i$ carries a learnable parameter $f_i \approx f(\mathbf{p}_i)$ that estimates the true SDF at that location. The collection of all Gaussians’ SDFs thus implicitly defines the scene’s surface as the zero-level set:
\begin{equation}
\mathcal{S} = \{ \mathbf{p} \in \mathbb{R}^3 \mid f(\mathbf{p}) = 0 \}.
\end{equation}

which we aim to reconstruct accurately. During training, we provide SDF supervision to guide $f_i$ toward this true signed distance field (details in Sec.~IV-C1), effectively encoding strong geometric priors into the model. The result is that geometry is no longer an afterthought: it is a core part of the representation, directly influencing how each primitive is treated in rendering. \par
To make use of these SDF values in our differentiable rendering process, we introduce an SDF-guided opacity modulation for the Gaussians. In standard Gaussian splatting (3DGS), each Gaussian has an opacity that determines how much it contributes along a ray, but this opacity is typically learned as an independent parameter or derived from image colors, offering only indirect geometric meaning. In DiGS, we compute each Gaussian’s opacity $\alpha_i$ as a function of its SDF value $f_i$. Specifically, we define a smooth kernel that peaks when the Gaussian lies on the surface ($f_i=0$) and decays as the Gaussian moves away from the surface:
\begin{equation}
\alpha_i =\exp \left(-\frac{f(p_i)^2}{\delta^2}\right).
\end{equation}

where $\delta$ is a learnable bandwidth parameter controlling the sharpness of the decay. This SDF-to-opacity mapping is a differentiable function that effectively locks each Gaussian to the surface: Gaussians with $f_i \approx 0$ (near the surface) will have high opacity (contribute strongly to rendering), whereas Gaussians with large $|f_i|$ (far from the surface either outside or inside) become nearly transparent. By plugging this $\alpha_i$ into the volume rendering, we ensure that only surface-aligned primitives are visibly rendered, and any Gaussian that drifts off the surface will automatically diminish in influence. The parameter $\delta$ can be learned or set to a small value so that the opacity drops off quickly with distance, yielding a tight approximation of an actual surface.\par
Crucially, our framework directly couples geometry and appearance via SDF-modulated opacity, unifying radiance and surface representations in a single branch. Each Gaussian encodes both geometry and color, ensuring consistency without the dual pipelines and enforcement mechanisms required by GSDF \cite{yu2024gsdf}. This tight integration simplifies optimization, avoids geometry–appearance misalignments, and yields more coherent, cross-view consistent reconstructions.

\subsection{Geometry-Guided Grid Growth} 
Even with multi-scale priors, an initial Gaussian grid may leave surfaces under-sampled, especially in weakly textured or SfM-sparse regions. To overcome this, DiGS employs a geometry-guided grid growth strategy that inserts new Gaussians during training based on depth and normal cues, ensuring accurate surface coverage. Unlike offline refinements, growth is integrated into end-to-end optimization, leveraging the LoD octree for placement and SDF/normal supervision for precision. The process follows three stages: (1) depth-normal estimation, (2) coarse-to-fine insertion, and (3) Progressive refinement with growth and pruning. Pseudocode is provided in Algorithm \ref{grid_growth} for reference. 
\begin{algorithm}[t]
\caption{Grid Growth Strategy} 
\label{grid_growth}
\begin{algorithmic}[1]
\Require 
    Training views $\{V_i\}$, 
    Depth/normal model,
    LoD octree, 
    $\theta_{\text{thresh}}$, $s_{\text{down}}$,
    Current iteration $t$
\Ensure
    Updated LoD structure

\If{$t = 5000$} \Comment{Trigger growth at iteration 5000}
    \For{each view $V_i \in \{V_i\}$}
        \State \textbf{Step 1: Generate depth/normal maps}
        \State $D_i, N_i \gets \text{DepthNormalEstimation}(V_i)$
        
        \State \textbf{Step 2: Filter observations}
        \For{each pixel $p$ in $D_i$}
            \State $\mathbf{v} \gets \text{ViewingDirection}(p)$
            \State $\mathbf{n} \gets N_i(p)$
            \If{$\arccos(|\mathbf{v} \cdot \mathbf{n}|) > \theta_{\text{thresh}}$}
                \State Discard $p$ \Comment{Remove grazing angles}
            \EndIf
        \EndFor
        
        \State \textbf{Step 3: Back-project point cloud}
        \State $\mathcal{P}_{\text{raw}} \gets \text{Backproject}(D_i, V_i.\text{camera})$
        \State $\mathcal{P}_{\text{filtered}} \gets \text{UniformDownsample}(\mathcal{P}_{\text{raw}}, s_{\text{down}})$
        
        \State \textbf{Step 4: Insert into LoD}
        \For{each point $\mathbf{p} \in \mathcal{P}_{\text{filtered}}$}
            \For{$\ell = 0$ \textbf{to} $L_{\max}$} \Comment{Coarse-to-fine insertion}
                \State $\text{Grid} \gets \text{FindGrid}(\mathbf{p}, \text{LoD}[\ell].\text{resolution})$
                \If{$\text{Grid not occupied}$}
                    \State Initialize grid with center at voxel.center
                    \State $\mathbf{n}_i \gets \text{Normal}(\mathbf{p}, N_i)$
                    \State $\mathbf{x}_i, \mathbf{y}_i \gets \text{OrthogonalTangents}(\mathbf{n}_p)$
                    \State $\mathbf{R} \gets [\mathbf{x_i}, \mathbf{y_i}, \mathbf{n}_i]$ \Comment{Local frame}
                    \State $\Sigma \gets \mathbf{R} \cdot \text{diag}(\sigma_t^2, \sigma_t^2, \sigma_n^2) \cdot \mathbf{R}^\top$
                    \For{$j = 1$ \textbf{to} $k$}
                        \State $\text{InitializeGaussianGrid}$
                        \State$(\text{position} = \text{Grid.center}, $
                        \State    $\text{covariance} = \Sigma)$
                        \State $\text{grid.AddGaussian}(G_j)$
                    \EndFor
                    \State $\text{LoD}[\ell].\text{AddGrid}(\text{grid})$
                    \State \textbf{break} \Comment{Insert at coarsest possible level}
                \EndIf
            \EndFor
        \EndFor
    \EndFor
\EndIf
\end{algorithmic}
\end{algorithm}
\subsubsection{Depth-Normal Estimation} 
We trigger the growth procedure at a certain training iteration (in our implementation, at $t=5000$ iterations). At that point, we leverage multi-view geometry from the input images to estimate dense depth and surface normal maps for each training view. For this purpose, we can employ a robust multi-view stereo technique or leverage the current state of the model itself. In our implementation, we found it effective to use a plane-guided depth estimation similar to the one in PGSR \cite{chen2024pgsr}: by enforcing local planarity and photometric consistency across views, we compute a dense depth map $D_i(u,v)$ and normal map $N_i(u,v)$ for each image $V_i$ (where $(u,v)$ are pixel coordinates). This process is akin to a one-pass stereo reconstruction from the images, and it benefits from the fact that by iteration 5000 our model’s poses and coarse shape are already reasonably aligned with the training images. \par
Once we have the depth $D_i$ and normal $N_i$ for a view $v$, we filter these observations to ensure we only add high-quality points. We iterate over each pixel $p=(u,v)$ and consider the depth value $d = D_i(u,v)$ with its corresponding normal $\mathbf{n} = N_i(u,v)$. We discard points that are likely erroneous or not robust: 
\begin{equation}
\text{if } \arccos(|\mathbf{v} \cdot \mathbf{n}|) > \theta_{\text{thresh}}, \quad \text{discard point}.
\end{equation}

if the normal $\mathbf{n}$ is not consistent with the viewing direction, or if the depth confidence is low. After filtering, we obtain a set of reliable surface points for each view. These points densely sample the scene surfaces as seen from that view. 

\subsubsection{Aggressive Coarse-to-Fine Insertion} With the collected 3D surface points from all views, we proceed to aggressively insert new Gaussians into our representation to cover these surfaces. The goal is to fill in any gaps in the current Gaussian set by using the depth-informed points as candidate locations for new primitives. A naive approach might add a Gaussian at the exact location of every point, but this could lead to an unmanageable number of primitives and a lot of redundancy. Instead, we insert points in a structured manner aligned with our LoD octree. 
For each candidate point $\mathbf{p}$ (with normal $\mathbf{n}$), we traverse the LoD hierarchy from coarse to fine and determine where to place Gaussians such that all levels are appropriately populated:\par
\textbf{LoD-Aware Placement:} Starting at the coarsest level $\ell=0$, we check if $\mathbf{p}$ falls inside an existing voxel at that level that already contains a Gaussian. If not (meaning the coarse grid had a hole at that location), we create a new Gaussian grid cell at level 0 covering $\mathbf{p}$ and initialize it with $k$ Gaussians (distributed by the same offset scheme as before). We orient the covariance of these new Gaussians according to the normal $\mathbf{n}$, i.e., we align each new Gaussian’s principal axis with $\mathbf{n}$ so that they lie roughly tangent to the surface. We also set the SDF values of these new Gaussians to $f\approx 0$ initially to integrate them smoothly into the model. \par
\begin{equation}
\boldsymbol{\Sigma} = \mathbf{R} 
\begin{bmatrix}
\sigma_t^2 & 0 & 0 \\
0 & \sigma_t^2 & 0 \\
0 & 0 & \sigma_n^2 \\
\end{bmatrix}
\mathbf{R}^\top,
\quad \text{where } \mathbf{R} = [\mathbf{a_i}, \mathbf{b_i}, \mathbf{n_i}].
\end{equation}
\begin{figure*}[t]
\centering
\includegraphics[width=6.9in]{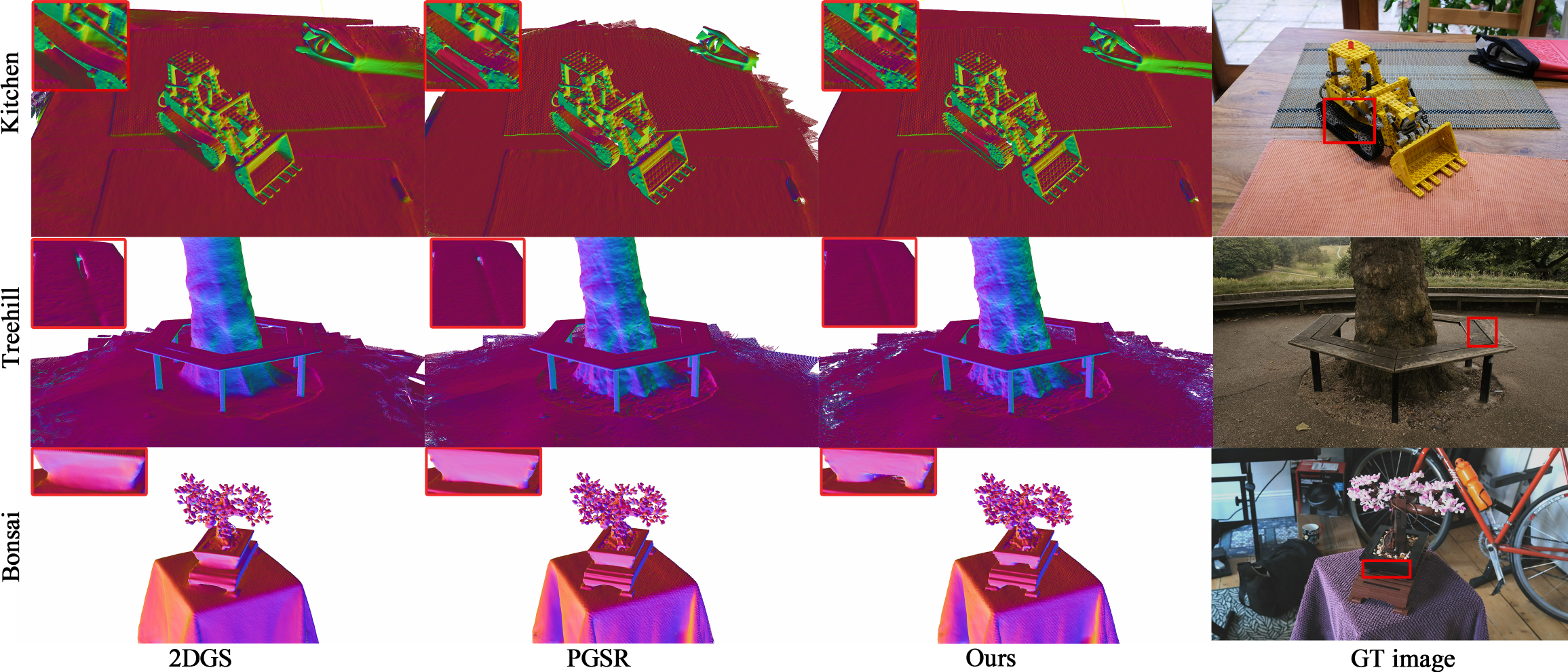}
\caption{The Qualitative Results on the Mip-NeRF 360 dataset \cite{barron2022mip}. Our method demonstrates superior reconstruction quality, particularly in complex geometric regions and fine structural details as highlighted by the red box. }
\label{360}
% \vspace{-0.2cm}
\end{figure*}

Here, $\sigma_t$ and $\sigma_n$ denote the Gaussian radii in the tangent and normal direction. We set $\sigma_t = 1.0$ and $\sigma_n= 0.1$ to limit the expansion of the Gaussian along the normal. This design allows each Gaussian to capture local surface curvature and orientation, laying a geometry-aware foundation for subsequent optimization stages.\par
We then proceed to level $\ell=1$: we find the child cell of the previously considered cell that contains $\mathbf{p}$, and similarly check if that child exists. If not, we insert a new cell at level 1 with Gaussians (again oriented by $\mathbf{n}$). We repeat this process for each level up to $\ell=L_{\max}$. In essence, each point $\mathbf{p}$ can spawn at most one new cell per level, ensuring that all scales have a representation of that surface point if it was missing. If at some level the cell already existed (because perhaps another nearby point already triggered its creation or it was present from initialization), we skip adding at that level to avoid duplicates.\par
\textbf{Avoiding Redundancy:} This coarse-to-fine insertion ensures hierarchical consistency: a newly added fine Gaussian will also have support from coarser levels. By design, we do not add a new Gaussian into an occupied voxel to avoid redundant representations. This strategy contrasts with a naive global insertion of points which could clump many Gaussians in an area already represented. By checking occupancy at each level, we keep the Gaussian distribution even and controlled. Moreover, inserting from coarse to fine allows the model to maintain approximate coverage at all scales, which is important for both rendering.\par
After processing all candidate points in this manner, we update the LoD octree data structure to include all newly created cells and Gaussians. The result is a significantly densified Gaussian representation that now explicitly covers surfaces that were previously missed or under-sampled. We call this an “aggressive” growth because it can substantially increase the number of Gaussians in one go. However, because we insert guided by actual geometry and enforce the octree structure, the growth is targeted and does not lead to gratuitous increase in regions that were already adequately modeled. \par
\subsubsection{Progressive Growth and Pruning} After the aggressive insertion step, we resume the end-to-end training of the model with the expanded set of Gaussians. To handle the sudden increase in parameters, we employ a progressive learning rate schedule: we temporarily boost the learning rate for the newly added Gaussians’ parameters while slightly reducing it for the pre-existing ones, ensuring the new Gaussians quickly adjust to integrate with the scene without disturbing the already learned structure. Over time, these learning rates are unified again as the new primitives become part of the model. This strategy helps maintain training stability during growth. \par
As training continues, the new Gaussians will have their SDF values and colors refined by the loss functions (Sec.~IV-C), and any initial errors in their placement will be corrected. Because we inserted Gaussians at all LoD levels for each point, the model can learn a consistent multi-scale representation of newly added regions: coarse new Gaussians ensure that even from distant viewpoints the region contributes, while fine ones add detail up close. The LoD coupling also means that our earlier interpolation scheme and SDF field remain continuous despite the insertions, we added features at new grid vertices, and our IDW interpolation naturally incorporates those when querying points in those regions.\par
To keep the model efficient, we also perform pruning of redundant or unhelpful Gaussians as training proceeds. In particular, if a coarse Gaussian’s region has been completely taken over by finer Gaussians, we can safely deactivate or remove that coarse Gaussian. This pruning prevents double-counting of contributions and focuses computation on the needed primitives. Similarly, if any Gaussian satisfies both low opacity ($\alpha < \tau_\alpha$) or the signed distance ($|\text{SDF}| > \tau_\text{sdf}$), we can remove it from the model to streamline the representation. This keeps the scene representation compact and clean, containing mostly surface-relevant Gaussians. \par

\subsection{Optimization and Loss Functions} 
With the representation and growth strategy defined, we now describe the optimization objectives that guide the training of DiGS. Our loss design follows the principle of treating geometry as a first-class objective, while also ensuring high-quality radiance reconstruction. We employ a set of complementary losses: (1) geometric losses that supervise the SDF values and enforce SDF properties, (2) a flattening loss that regularizes Gaussian shape along surfaces, (3) an appearance loss for RGB reconstruction, and (4) a combined objective that balances all terms. Each loss term plays a specific role, and importantly, they act in concert on our unified pipeline. This means, for example, that improving geometric alignment via SDF losses also immediately benefits the rendering, and vice versa. We detail each category of loss below. 
\begin{table*}[t]
\caption{Quantitative Results of Chamfer Distance (mm) on DTU Dataset. Our method outperforms existing 3D GS and NeRF-based approaches, achieving state-of-the-art accurate reconstruction quality on the DTU dataset. ”Red”, ”Orange” and ”Yellow” denote the best, second-best, and third-best results.}
\centering
\renewcommand\arraystretch{1.0}
\resizebox{2.0\columnwidth}{!}{
\begin{tabular}{llllllllllllllllll}
\toprule
\textbf{Method} & \textbf{24} & \textbf{37} & \textbf{40} & \textbf{55} & \textbf{63} & \textbf{65} & \textbf{69} & \textbf{83} & \textbf{97} & \textbf{105} & \textbf{106} & \textbf{110} & \textbf{114} & \textbf{118} & \textbf{122} & \textbf{Mean} &\textbf{Time}\\ 
\midrule
VolSDF &1.14 &1.26 &0.81 &0.49 &1.25& 0.70 &0.72 &1.29 &1.18& 0.70 &0.66& 1.08 &0.42 &0.61 &0.55&0.86&  \textgreater 12h \\
NeuS & 1.00 &1.37 &0.93 &0.43 &1.10 &0.65 &0.57 &1.48 &1.09 &0.83 &0.52 &1.20& 0.35 &0.49 &0.54&0.84&  \textgreater 12h \\
Neuralangelo & \cellcolor{orange!40}0.37 &\cellcolor{yellow!40}0.72& \cellcolor{yellow!40}0.35 &\cellcolor{yellow!40}0.35 &\cellcolor{yellow!40}0.87 &\cellcolor{orange!40}0.54 &\cellcolor{orange!40}0.53 &1.29 &\cellcolor{yellow!40}0.97 &0.73& \cellcolor{orange!40}0.47 &\cellcolor{yellow!40}0.74 &\cellcolor{yellow!40}0.32 &\cellcolor{yellow!40}0.41& \cellcolor{yellow!40}0.43&  \cellcolor{yellow!40}0.61 &  \textgreater 128h\\
\midrule

SuGaR &1.47 &1.33 &1.13 &0.61 &2.25 &1.71 &1.15 &1.63 &1.62 &1.07 &0.79 &2.45& 0.98 &0.88 &0.79&
1.33&   1h  \\
GSDF &0.59 &0.94 &0.46 &0.38 &1.30 &0.77 &0.73 &1.59 &1.29 &0.76 &0.59 &1.22 &0.38 &0.52 &0.51&0.80&  \textgreater 2h\\
Scaffold-GS & 7.23& 6.23 &6.48 &7.44 &8.17 &4.27 &5.78 &5.45 &6.57 &6.36 &5.05 &5.95 &6.32 &5.62 &2.90&4.63&  0.8h\\
Octree-GS & 4.3& 3,45 &4,56 &3.45 &5.67 &2.345&3.45 &4.45 &3.32 &3.35 &2.45 &4.56&4.45 &3.45 &4.34&4.23&  0.7h\\
2DGS & 0.46 &0.91 &0.39 &0.39 &1.01 &0.83 &0.81 &1.36 &1.27 &0.76 &0.70& 1.40 &0.40 &0.76 &0.52&
0.80  &  0.32h\\
GOF & 0.50 &0.82 &0.37 &0.37 &1.12 &0.74 &0.73 &\cellcolor{yellow!40}1.18 &1.29 &\cellcolor{orange!40}0.68 &0.77 &0.90 &0.42 &0.66 &0.49&
0.74&  2h\\
PGSR &\cellcolor{red!40}\textbf{0.34} &\cellcolor{orange!40}0.58 &\cellcolor{orange!40}0.29 &\cellcolor{red!40}\textbf{0.29} &\cellcolor{orange!40}\textbf{0.78} &\cellcolor{yellow!40}0.58& \cellcolor{yellow!40}0.54& \cellcolor{red!40}\textbf{1.01} &\cellcolor{orange!40}0.73 &\cellcolor{orange!40}\textbf{0.51} &\cellcolor{orange!40}0.49 &\cellcolor{orange!40}0.69 &\cellcolor{orange!40}0.31 &\cellcolor{orange!40}0.37 &\cellcolor{orange!40}0.38&
\cellcolor{orange!40}0.53 &  1h\\
Ours & \cellcolor{yellow!40}0.45	&\cellcolor{red!40}\textbf{0.44}	&\cellcolor{red!40}\textbf{0.25}	&\cellcolor{orange!40}0.33	&\cellcolor{red!40}\textbf{0.69}	&\cellcolor{red!40}\textbf{0.49}	&\cellcolor{red!40}\textbf{0.46}	&\cellcolor{orange!40}0.92	&\cellcolor{red!40}\textbf{0.57}	&\cellcolor{red!40}\textbf{0.47}	&\cellcolor{red!40}\textbf{0.43}	&\cellcolor{red!40}\textbf{0.46}	&\cellcolor{red!40}\textbf{0.28}	&\cellcolor{red!40}\textbf{0.35}	&\cellcolor{red!40}\textbf{0.32}	&\cellcolor{red!40}\textbf{0.46}&  1.2h\\ 
\bottomrule
\end{tabular}
}
\label{dtu_results}

\end{table*}
\begin{table}[t]
\caption{Quantitative Results of Reconstruction on TnT Dataset\cite{Knapitsch2017}.}
\centering
\renewcommand\arraystretch{1.0}
\resizebox{1.0\columnwidth}{!}{
\begin{tabular}{lccccccc}
\toprule
\multicolumn{1}{c}{Scene} & NeuS & Geo-Neus & Neuralangelo & 2D GS & GOF & PGSR & Ours \\
\midrule
Barn         & 0.29 & 0.33 & \cellcolor{red!40}\textbf{0.70} & 0.36 & 0.44 & \cellcolor{yellow!40}0.66 & \cellcolor{orange!40}0.68\\
Caterpillar  & 0.29 & 0.26 & 0.36 & 0.23 & \cellcolor{orange!40}0.41 & \cellcolor{orange!40}0.41 & \cellcolor{red!40}\textbf{0.42}\\
Courthouse   & 0.17 & 0.12 & \cellcolor{red!40}\textbf{0.28} & 0.13 & \cellcolor{red!40}\textbf{0.28} & 0.21 & \cellcolor{orange!40}0.25\\
Ignatius     & 0.83 & 0.72 & \cellcolor{red!40}\textbf{0.89} & 0.44 & 0.68 & \cellcolor{yellow!40}0.80 & \cellcolor{orange!40}0.83 \\
Meetingroom  & 0.24 & 0.20 & 0.32 & 0.16 & 0.28 & \cellcolor{red!40}\textbf{0.29} & \cellcolor{orange!40}0.28\\
Truck        & 0.45 & 0.45 & 0.46 & 0.26 & \cellcolor{yellow!40}0.58 & \cellcolor{orange!40}0.60 & \cellcolor{red!40}\textbf{0.61} \\
\midrule
Mean         & 0.38 & 0.35 & \cellcolor{orange!40}0.50 & 0.30 & 0.46 & \cellcolor{orange!40}0.50 & \cellcolor{red!40}\textbf{0.51} \\
% Time         & >24h & >24h & >128h & 34.2m & 2h & 1.2h \\
\bottomrule
\end{tabular}
}
% \vspace{0.1cm}

% \vspace{-0.5cm}
\label{tnt}
\end{table}

\begin{table}[t]
\caption{Performance Comparison of Novel View Synthesis on Mip-NeRF 360 dataset \cite{barron2022mip}. Our method has a strong performance.}
\centering
\renewcommand\arraystretch{0.8}
\resizebox{1.0\columnwidth}{!}{
\begin{tabular}{llllllllll}
\toprule
& \multicolumn{3}{c}{\textbf{Indoor Scenes}} & \multicolumn{3}{c}{\textbf{Outdoor Scenes}} & \multicolumn{3}{c}{\textbf{All Scenes}} \\
\cmidrule(lr){2-4} \cmidrule(lr){5-7} \cmidrule(lr){8-10}
\textbf{Method} & PSNR↑ & SSIM↑ & LPIPS↓ & PSNR↑ & SSIM↑ & LPIPS↓ & PSNR↑ & SSIM↑ & LPIPS↓ \\
\midrule
% \multicolumn{10}{l}{\textit{NeRF-based Methods}} \\
NeRF & 26.84 & 0.790 & 0.370 & 21.46 & 0.458 & 0.515 & 24.15 & 0.624 & 0.443 \\
Deep Blending & 26.40 & 0.844 & 0.261 & 21.54 & 0.524 & 0.364 & 23.97 & 0.684 & 0.313 \\
Instant-NGP & 29.15 & 0.880 & 0.216 & 22.90 & 0.566 & 0.371 & 26.03 & 0.723 & 0.294 \\
Mip-NeRF360 &  \cellcolor{red!40}\textbf{31.72} & 0.917 & 0.180 & 24.47 & 0.691 & 0.283 & 28.10 & 0.804 & 0.232 \\
NeuS  & 25.10 & 0.789 & 0.319 & 21.93 & 0.629 & 0.600 & 23.74 & 0.720 & 0.439 \\
\midrule
% \multicolumn{10}{l}{\textit{GS-based Methods}} \\
3DGS  & \cellcolor{orange!40}30.99 & 0.926 & 0.199 & 24.24 & 0.705 & 0.283 & 27.24 & 0.803 & 0.246 \\
2DGS  &30.39	&0.923	&0.183	&24.33	&0.709	&0.284	&27.03	&0.804	&0.239 \\
SuGaR  & 29.44 & 0.911 & 0.216 & 22.76 & 0.631 & 0.349 & 26.10 & 0.771 & 0.283 \\
GOF  &\cellcolor{yellow!40}30.80	&\cellcolor{orange!40}0.928	&\cellcolor{yellow!40}0.167	& \cellcolor{red!40}\textbf{24.76}	& \cellcolor{red!40}\textbf{0.742}	&\cellcolor{orange!40}0.225	& \cellcolor{red!40}\textbf{27.78}	& \cellcolor{red!40}\textbf{0.835}	&\cellcolor{orange!40}0.196 \\
PGSR  &30.41	& \cellcolor{red!40}\textbf{0.930}	& \cellcolor{red!40}\textbf{0.161}	&\cellcolor{yellow!40}24.45	&\cellcolor{yellow!40}0.730	& \cellcolor{red!40}\textbf{0.224}	&\cellcolor{yellow!40}27.43	&\cellcolor{yellow!40}0.830	& \cellcolor{red!40}\textbf{0.193} \\
Ours &30.60	&\cellcolor{yellow!40}0.927	&\cellcolor{yellow!40}0.171	&\cellcolor{orange!40}24.73	&\cellcolor{orange!40}0.739	&\cellcolor{yellow!40}0.254	&\cellcolor{orange!40}27.67	&\cellcolor{orange!40}0.833	&\cellcolor{yellow!40}0.212 \\
\bottomrule
\end{tabular}
}
\label{r_360}
% \vspace{-0.9cm}
\end{table}
\subsubsection{Geometric Losses} 
To supervise the SDF values attached to Gaussians, we design losses that encourage those values to reflect true surface distances and to satisfy the mathematical properties of an SDF. First, we introduce an SDF center loss $L_{\text{SDF}}$ that penalizes the absolute SDF value at any Gaussian which is expected to lie on the surface. Concretely, for each Gaussian $i$ that corresponds to a measured surface point, we add a term $|f(p_i)- 0|$ to the loss. We determine which Gaussians should lie on the surface. Then:
\begin{equation}
\mathcal{L}_{\text{SDF-Center}} = \sum_{i} |f(p_i)|^2.
\end{equation}

driving those $f(p_i)$ to zero. Next, we impose the Eikonal loss \cite{gropp2020implicit} to ensure the learned SDF behaves like a true signed Distance Field in the continuous space. The Eikonal term enforces that the gradient magnitude of $f(\mathbf{p})$ is 1 everywhere (which is a defining property of distance fields). We then add:
\begin{equation}
\mathcal{L}_{\text{Eikonal}} = \mathbb{E}_{\mathbf{pix} \sim \mathcal{X}} \left( \| \nabla f(p) \|_2 - 1 \right)^2.
\end{equation}

which encourages $|\nabla f|=1$. This regularizes the SDF field to prevent pathological solutions (e.g., constant or zero everywhere) and promotes smoothness in geometry. It also couples with the SDF center loss: having $f=0$ at surface points and unit gradient means that nearby points will get correct sign distances. Overall, these geometric losses provide strong supervision of the SDF compared to prior methods: for example, PGSR \cite{chen2024pgsr} used normal consistency losses indirectly, whereas we directly learn the SDF which inherently encodes surface normals (via $\nabla f$) and distances. By integrating this into our pipeline, we ensure multi-view geometric consistency is achieved through the SDF representation itself, rather than requiring separate multi-view consistency losses. This simplifies the training and yields a more interpretable geometric model as a byproduct. The overall SDF consistency can be geometrically regularized to:
\begin{equation}
\mathcal{L}_{\text{SDF}} =  \lambda_4 \mathcal{L}_{\text{sdf-center}} + \lambda_5 \mathcal{L}_{\text{Eikonal}}.
\end{equation}

$\lambda_4$ and $\lambda_5$ are the parameters of the corresponding loss.
\subsubsection{Flattening Loss for Gaussian Covariance} 
Each Gaussian is an anisotropic ellipsoid parameterized by covariance \(\mathbf{\Sigma}_i = \mathbf{R}_i \mathbf{S}_i \mathbf{S}_i^\top \mathbf{R}_i^\top\), where \(\mathbf{R}_i \in \mathbb{R}^{3\times3}\) is a rotation matrix and \(\mathbf{S}_i = \mathrm{diag}(s_{i1}, s_{i2}, s_{i3})\) scales along principal axes. Following \cite{huang20242d}, the smallest scale direction defines the local surface normal \(\mathbf{n}_i\). Minimizing this smallest scale encourages flattening along \(\mathbf{n}_i\), promoting compact, planar Gaussians:
\[
\mathcal{L}_s = \lambda_6 \sum_i \min(s_{4}, s_{5}, s_{6}).
\] 
\subsubsection{RGB Reconstruction Loss} For appearance, we use an RGB reconstruction loss to train the color output of our model. We render each training image via our differentiable splatting (with SDF-based opacities) and compare to the ground truth image. We denote by $C_i(u,v)$ the color predicted by our model for pixel $(u,v)$ in view $V_i$, and $C_i^{GT}(u,v)$ the corresponding ground truth pixel color. We then define:
\begin{equation}
\mathcal{L}_{\text{rgb}} = (1-\lambda_7) \|\tilde{I} - I_i\|_1 + \lambda_7 \cdot \text{SSIM}(\tilde{I}, I_i).
\label{eq:rgb_loss}
\end{equation}

where $\tilde{I}$ denotes the rendered image, $\tilde{I}$ denotes the ground truth image and $\lambda_1$ controls the trade-off.
\subsubsection{Final Objective} The total loss is a weighted sum of the above terms:
\begin{equation}
\mathcal{L}=\mathcal{L}_{\mathrm{rgb}}+\mathcal{L}_s+\mathcal{L}_{\mathrm{geo}}+\mathcal{L}_{\mathrm{SDF}} .
\end{equation}

\section{Experiment}

\subsection{Experimental Setup}
\begin{figure*}[!t]
\centering
\includegraphics[width=6.7in]{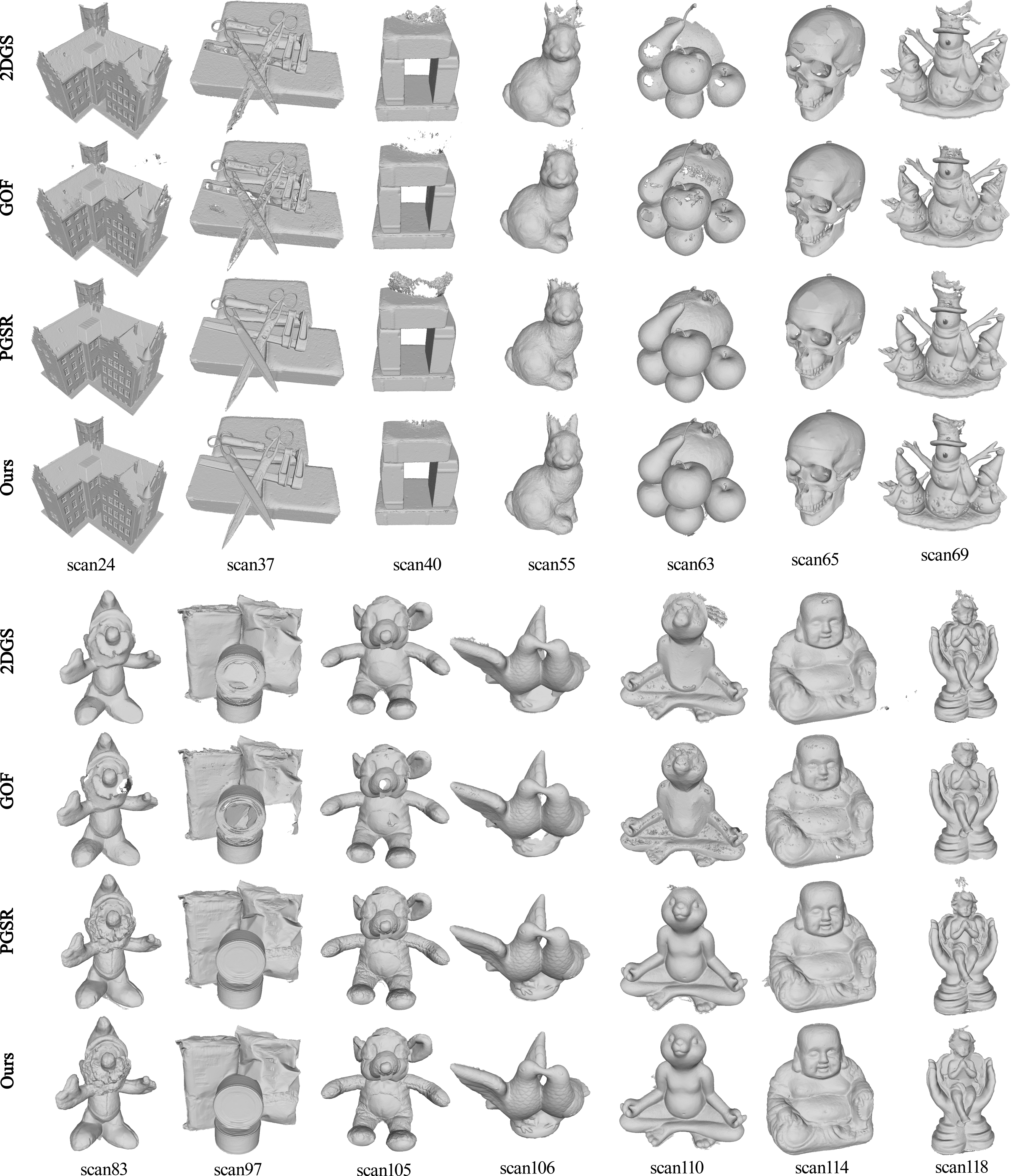}
\caption{The Qualitative Results on DTU dataset. Our method has more complete and accurate reconstruction results in scenes scan40, scan55, and scan69, and better reconstruction details in scan37, scan65, and scan110. }
\label{dtu2}
\end{figure*}
\textbf{Datasets and Baselines.} We evaluate our method on three diverse datasets: DTU \cite{aanaes2016large}, Mip-NeRF 360 \cite{barron2022mip}, TnT \cite{Knapitsch2017}. These datasets cover a range of 3D scene complexities, providing a comprehensive benchmark for reconstruction accuracy and visual fidelity. For comparison, we select a set of state-of-the-art reconstruction methods: NeRF \cite{mildenhall2021nerf}, NeuS \cite{wang2021neus}, Neuralangelo \cite{li2023neuralangelo}, Deep Blending \cite{hedman2018deep}, INGP \cite{muller2022instant}, 3DGS \cite{kerbl20233d}, Scaffold-gs \cite{lu2024scaffold}, Octree-gs \cite{ren2024octree}, SuGar \cite{guedon2024sugar}, and PGSR \cite{chen2024pgsr}, each representing a notable approach in 3D scene surface reconstruction.

\textbf{Evaluation Metrics.} We assess the performance of our method using several standard reconstruction and rendering metrics: Chamfer Distance, F1-score, along with image quality metrics including PSNR, SSIM, and LPIPS. These metrics allow for a holistic evaluation of both geometric fidelity and photorealistic rendering quality. Mesh reconstruction is achieved using a TSDF-based extraction method \cite{chen2024pgsr}, which ensures consistency in the surface.

\textbf{Implementation Details.} The optimization process incorporates multiple loss components, including L1 loss, SSIM loss, image photometric loss, Gaussian flattening loss, single-view loss, and multi-view loss. The corresponding weight factors, $\lambda_4$ to $\lambda_7$, are set to 0.01, 0.01, 100, and 0.05, respectively. These weights are carefully tuned to balance the contributions of each loss term, ensuring optimal convergence and faithful reconstruction throughout the training process.

Our training follows the schedule outlined in 3D Gaussian Splatting, with a total of 30,000 epochs. The densification process begins at epoch 1500 and concludes at epoch 15,000, facilitating gradual scene refinement. To enhance geometric accuracy in sparse regions, the depth-aggressive growth strategy is activated at epoch 5,000. We employ a progressive learning rate schedule for neural network components, gradually increasing their learning capacity over time, while maintaining a fixed learning rate for the grid positions to preserve spatial stability throughout training.

All experiments are conducted on a single NVIDIA RTX 3090 GPU. For large-scale datasets that exceed the GPU memory capacity, data loading is offloaded to the CPU. This strategy mitigates GPU memory pressure, ensuring stable training throughput and the ability to handle larger datasets without compromising performance.

\begin{figure*}[!t]
\centering
\includegraphics[width=7.3in]{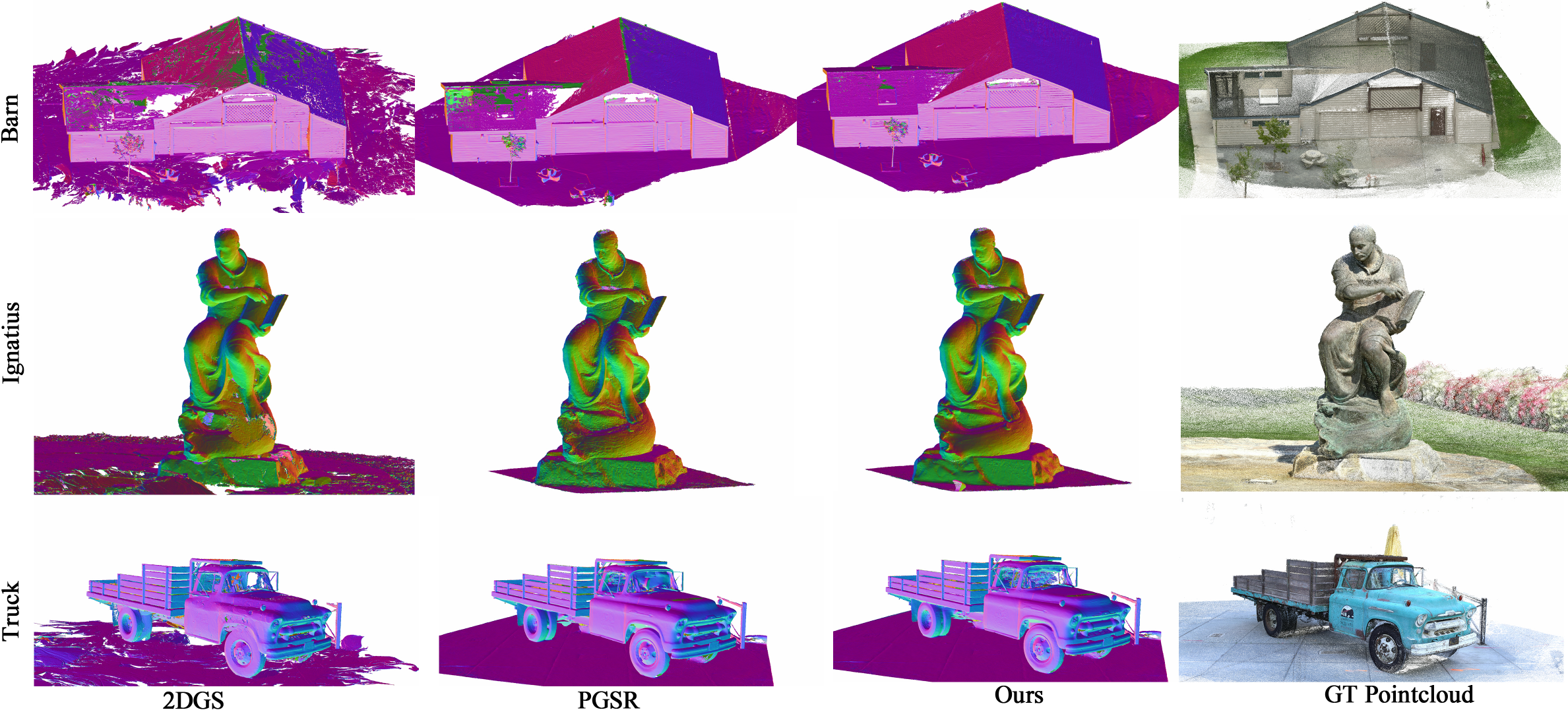}
\caption{The Qualitative Comparison of Reconstruction Results on the Truck, Ignatius, and Barn scenes from the TnT Dataset \cite{Knapitsch2017}. Our method demonstrates superior fidelity in preserving geometric details and structural integrity. Notably, our approach achieves sharper edges, and more consistent surface continuity.}
\label{recon_tnt}
\end{figure*}
% \vspace{-0.5em}
% \subsection{Reconstruction}
% On the DTU dataset, our method achieves state-of-the-art surface reconstruction accuracy, outperforming existing 3D Gaussian Splatting (3DGS) and neural implicit methods (Table~\ref{dtu_results}). The key advantage lies in our SDF-constrained 3D Gaussians, which enhance geometric fidelity by tightly binding the primitives to surfaces. Additionally, our normal-aligned, depth-guided Aggressive Gaussian Grid technique adaptively distributes Gaussian primitives to ensure full coverage, effectively addressing the incomplete growth problem seen in methods like GOF. Further refined by LOD-driven multi-scale optimization, our approach balances high-detail preservation with computational efficiency. Figures~\ref{dtu2} demonstrate these improvements, showcasing more complete and geometrically accurate reconstructions compared to prior work.

\subsection{Reconstruction}

\textbf{Accuracy.}  
We evaluate reconstruction accuracy on both DTU \cite{aanaes2016large} and Tanks \& Temples (TnT). Table~\ref{dtu_results} reports Chamfer Distance on DTU, where our method achieves the lowest error among all compared approaches, clearly outperforming Gaussian-based 3DGS \cite{kerbl20233d} and neural implicit baselines such as NeRF \cite{mildenhall2021nerf} and NeuS \cite{wang2021neus}. Table~\ref{tnt} further presents F1 scores on TnT, which jointly capture precision and recall under a strict distance threshold. Our method consistently achieves the highest average F1, indicating both accurate alignment to ground truth geometry and reliable coverage of fine details. These results demonstrate that explicitly embedding signed distance supervision into Gaussian primitives significantly improves geometric fidelity across benchmarks.  

Qualitative results corroborate the quantitative improvements. As shown in Figures~ \ref{360}, \ref{dtu2},and \ref{recon_tnt}, our reconstructions exhibit sharper boundaries, cleaner topology, and better preservation of thin structures compared to competing methods. In contrast, baselines such as GOF \cite{yu2024gaussian} often display surface drift, misalignments, or incomplete recovery, particularly in regions with fine-scale details or weak textures. Our method also generalizes robustly to real-world Mip-NeRF 360 scenes, where it produces geometrically faithful surfaces despite complex illumination and scale variations.  

\textbf{Completeness.}  
Beyond accuracy, we assess the completeness of reconstructed surfaces, which is particularly challenging in occluded or low-texture regions. Prior Gaussian-based pipelines often rely on appearance-driven growth strategies, resulting in uneven Gaussian distributions and noticeable holes. In contrast, our geometry-guided aggressive grid growth adaptively expands Gaussians along surface-consistent directions, ensuring dense coverage and robust scene reconstruction.  

Figures~\ref{360} and \ref{recon_tnt} highlight this advantage: our method yields reconstructions with more uniform Gaussian placement and significantly fewer missing regions, even under severe occlusions or texture sparsity. Competing methods frequently fail to recover such areas, leaving gaps or fragmented structures, while our approach produces coherent and watertight models. These results confirm that the proposed growth mechanism plays a critical role in achieving visually complete reconstructions, complementing the quantitative accuracy improvements.  
\begin{figure*}[!t]
\centering
\includegraphics[width=6.8in]{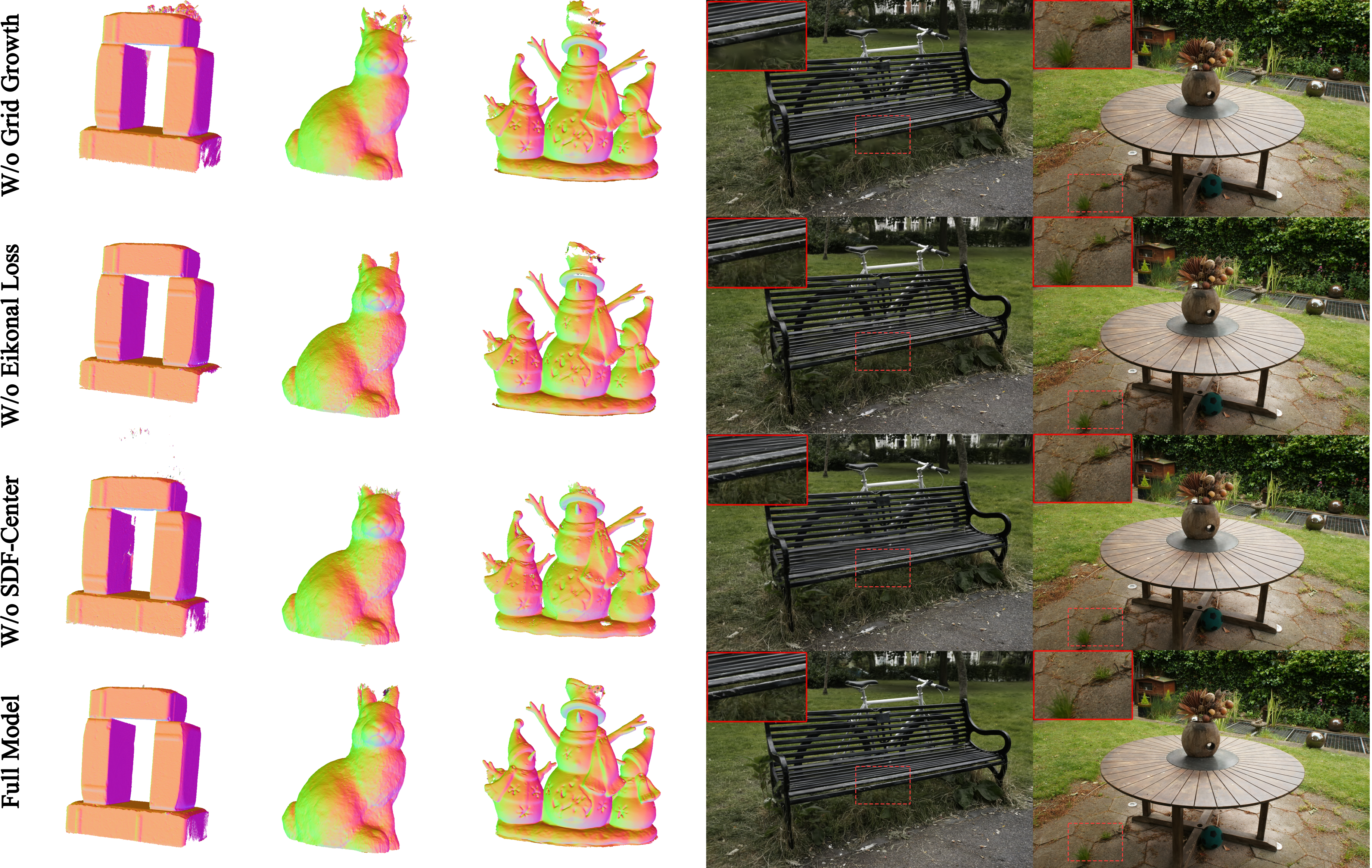}
\caption{Qualitative ablation on DTU \cite{aanaes2016large} and Mip-NeRF360 \cite{barron2022mip}. Rows from top to bottom show the model without geometry-guided grid growth, without Eikonal loss, without SDF-center loss, and the full model. Left block shows reconstructed geometry with surface normals on DTU objects. Right block shows novel view synthesis results on Mip-NeRF360 with zoom-in insets. Removing any component degrades edge sharpness, normal consistency, and coverage, while the full model recovers thin structures and fills missing regions.}
\label{fig:ablation}
\end{figure*}

\begin{figure}[t]
\centering
\includegraphics[width=0.5\textwidth]{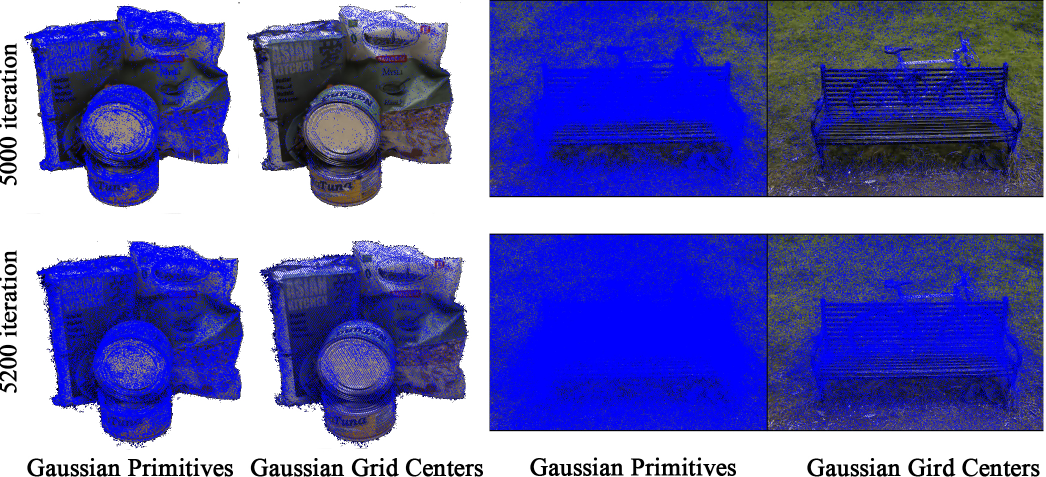}
\caption{The Process of Aggressive Grid Growth strategy. At 5000 iterations, the Grid Growth strategy is executed. }
\label{fig:grow}
% \vspace{-0.6cm}
\end{figure}

\subsection{Novel View Synthesis}

For the evaluation of novel view synthesis quality, we adopt the rigorous evaluation protocol established by 3DGS and conduct a comprehensive assessment on the Mip-NeRF360 \cite{barron2022mip} benchmark dataset. This allows us to thoroughly evaluate the proposed method's performance in generating novel views from sparse input perspectives. To ensure a balanced and fair comparison, we benchmark our method against two categories of state-of-the-art (SOTA) approaches: (1) methods that have demonstrated exceptional performance in novel view synthesis tasks and (2) surface reconstruction techniques that share similar technical principles with our framework.

As detailed in Table \ref{r_360}, our approach not only excels in high-precision surface geometry reconstruction but also achieves remarkable performance in terms of visual quality metrics for novel view synthesis. This includes improvements in image fidelity, detail retention, and depth consistency when viewed from unseen viewpoints. The results clearly highlight that our method balances geometric accuracy with photorealistic rendering quality, offering a more robust solution for novel view synthesis compared to existing techniques. These advancements are underpinned by the synergy between our SDF-constrained 3D Gaussian representation and adaptive grid growth strategy, which together enhance both geometric consistency and visual realism.

\begin{table*}[t]
\footnotesize
\caption{Quantitative ablation on DTU \cite{aanaes2016large} and Mip-NeRF360 \cite{barron2022mip}. The full model achieves the best scores on both datasets, confirming the complementary effects of SDF anchoring, Eikonal regularization, and geometry-guided aggressive grid growth.}
\centering
\renewcommand\arraystretch{0.7}
\resizebox{1.8\columnwidth}{!}{
\begin{tabular}{llllllll}
\toprule
& \multicolumn{4}{c}{\textbf{DTU}}  & \multicolumn{3}{c}{\textbf{Mip-NeRF360}} \\
\cmidrule(lr){2-5} \cmidrule(lr){6-8} 
\textbf{Method} & CD↓ & PSNR↑ & SSIM↑ & LPIPS↓ & PSNR↑ & SSIM↑ & LPIPS↓ \\
\midrule
% \multicolumn{10}{l}{\textit{NeRF-based Methods}} \\
W/o SDF-Center constraint & 0.49 & 34.65& 0.74 & 0.370 &27.43&0.828&0.217  \\
W/o Eikonal Regularization & 0.51 & 34.18 & 0.68 & 0.361 &27.33&0.827&0.220 \\
W/o Aggressive Grid Growth & 0.50 & 34.13  & 0.70& 0.317&27.45&0.826&0.227 \\
% Fixed resolution & 0.52 & 34.7  & 0.78& 0.286  \\
Full Model  & 0.46 & 35.63 & 0.81 & 0.219 &27.67	&0.833	&0.212   \\
\bottomrule
\end{tabular}
}

\label{tab:ablation}
% \vspace{-0.9cm}
\end{table*}

\subsection{Ablation Study}

We evaluate three principal design choices of our method: \emph{SDF-constrained 3D Gaussians} enforced by an SDF-center loss, \emph{Eikonal regularization} for gradient and normal consistency, and \emph{geometry-guided aggressive grid growth} driven by depth and normal cues. Quantitative results on DTU and Mip-NeRF360 are reported in Table~\ref{tab:ablation}. Qualitative comparisons are shown in Fig.~\ref{fig:ablation}, and the growth schedule is illustrated in Fig.~\ref{fig:grow}. Across both datasets, removing any single component consistently degrades geometric accuracy and perceptual quality, which indicates complementary effects among the proposed modules.

\paragraph{SDF-constrained 3D Gaussians representation}
Binding each Gaussian to the SDF zero level set through the center prior provides geometric anchoring that suppresses surface drift and floating artifacts. Without this loss, CD on DTU increases from 0.46 to 0.49, PSNR decreases from 35.63 to 34.65, and SSIM drops from 0.81 to 0.74. On Mip-NeRF360, PSNR declines from 27.67 to 27.43, SSIM from 0.833 to 0.828, and LPIPS rises from 0.212 to 0.217. Normal visualizations in Fig.~\ref{fig:ablation} reveal softened edges and small-scale misalignments in high-curvature regions. These observations show that SDF anchoring turns Gaussians into surface-attached primitives rather than view-dependent volumetric proxies.

\paragraph{Eikonal regularization}
The Eikonal term enforces unit SDF gradients and yields well-behaved normals and a thin, stable level set. Removing this regularization leads to over-smoothed geometry. On DTU, CD rises to 0.51, SSIM falls to 0.68, and PSNR decreases by 1.45\,dB from 35.63 to 34.18. On Mip-NeRF360, PSNR reduces to 27.33, SSIM to 0.827, and LPIPS increases to 0.220. Visual evidence in Fig.~\ref{fig:ablation} shows loss of crisp boundaries on the bench slats and the tabletop rim together with locally inconsistent normals. These results highlight the role of the Eikonal term in preserving fine detail and enforcing a coherent surface.

\paragraph{Geometry-guided aggressive grid growth}
We introduce a one-shot aggressive expansion of the Gaussian grid scheduled at 5k iterations and guided by depth residuals and normal disagreement. In contrast to uniform or heuristic upsampling, this policy allocates new primitives only where geometry is under-explained, which rapidly covers unmodeled regions while avoiding redundant capacity in flat areas. Disabling this module reduces DTU SSIM from 0.81 to 0.70 and increases LPIPS from 0.219 to 0.317. On Mip-NeRF360, PSNR drops from 27.67 to 27.45, SSIM from 0.833 to 0.826, and LPIPS rises to 0.227. Fig.~\ref{fig:ablation} shows incomplete coverage and gaps along thin structures such as bench slats and in low-texture ground regions when the growth policy is removed. The data indicate that deciding where and when to grow matters as much as the total amount of growth.

% % \paragraph{Full model and synergy}
% Combining all components yields the strongest results on both benchmarks. On DTU, the full model achieves CD 0.46, PSNR 35.63, SSIM 0.81, and LPIPS 0.219. On Mip-NeRF360, it reaches PSNR 27.67, SSIM 0.833, and LPIPS 0.212. Relative to the ablated variants on Mip-NeRF360, the full model improves PSNR by 0.22–0.34\,dB and reduces LPIPS by 0.005–0.015 absolute. On DTU, it lowers LPIPS from 0.317 to 0.219 and raises SSIM from as low as 0.68 to 0.81. The consistent improvements support three claims: SDF anchoring converts Gaussians into faithful surface primitives, Eikonal regularization sharpens and stabilizes the level set, and geometry-guided aggressive growth delivers targeted capacity exactly where reconstruction is most challenging. Together, these innovations enable high geometric fidelity and photorealism with efficient training dynamics.

\section{Conclusion}

We presented DiGS, a unified framework that embeds signed distance supervision directly into Gaussian primitives and integrates a geometry-guided grid growth strategy under a multi-scale hierarchy, thereby transforming Gaussian splatting into a geometry-preserving paradigm. By tightly coupling geometry and appearance within a single optimization process, DiGS achieves accurate and complete surface reconstruction while retaining high-fidelity rendering. Extensive experiments validate that this joint design yields sharp structural details, robust surface consistency, and efficient scalability across diverse datasets. We envision future extensions of DiGS toward explicit mesh integration, dynamic scene modeling, and data compression, further broadening its applicability in real-time, geometry-aware 3D reconstruction.

% \bibliographystyle{IEEEtran}
% \bibliography{reference}
% \clearpage         % 或者 \newpage，确保附录从新页开始
% \appendix  

% \input{references}

\bibliographystyle{IEEEtran}
\bibliography{reference.bib}

\end{document}